\documentclass[runningheads]{llncs}
\usepackage[T1]{fontenc}
\usepackage{cite} 
\usepackage{booktabs}

\usepackage{multirow}
\usepackage{amsmath,amssymb,amsfonts}
\usepackage{array}                              
\newcolumntype{P}[1]{>{\centering\arraybackslash}p{#1}}
\usepackage{algorithmic}
\usepackage{graphicx}
\usepackage{pifont}           
\usepackage{xspace}  
\usepackage{xcolor}
\usepackage[colorlinks=true,linkcolor=red,citecolor=blue,urlcolor=blue]{hyperref}
\newcommand{\method}{MRI2Rep\xspace} 
\usepackage{textcomp}
\usepackage{graphicx,verbatim}
\usepackage{comment}
\begin{document}
\title{MRI2Rep: Autoregressive Structured Report Generation for 3D Liver MRI}
\titlerunning{MRI2Rep: Structured Report Generation for 3D Liver MRI}
%
\author{Xinran Li\inst{1} \and
Junlin Yang\inst{2} \and
Annabella Shewarega\inst{2} \and
Zongwei Zhou\inst{4, 5} \and
Julius Chapiro\inst{2} \and
James S. Duncan\inst{1, 2, 3} \and
Lawrence H. Staib\inst{1, 2, 3}\thanks{Corresponding author.}}
\authorrunning{X. Li et al.}

\institute{Department of Biomedical Engineering \and
    Department of Radiology \& Biomedical Imaging \and
    Department of Electrical Engineering \\
    Yale University, New Haven, CT 06520, USA
 \and
Johns Hopkins University, Baltimore, MD 21218, USA \\
 \and
Johns Hopkins Medicine, Baltimore, MD 21287, USA \\
\email{lawrence.staib@yale.edu}}

  
\maketitle              
\begin{abstract}
Manual reporting of 3D MRI studies is time-consuming, yet end-to-end, structured report generation for 3D liver MRI remains underexplored due to volumetric complexity and scarce paired data. We propose MRI2Rep, an autoregressive framework for liver MRI report generation. From 3{,}929 real-world MRI--report pairs acquired over a 10-year single-institution cohort, a Report-to-Label Canonicalization (RLC) module converts the free-text reports into structured, closed-vocabulary diagnostic sequences without lesion-level annotations. On a held-out test set, MRI2Rep achieves 76.0\% case-level sensitivity, 29.4\% lesion-level F1 (vs.\ $\leq$8.3\% for adapted medical vision--language baselines), and 82.4\% liver-level accuracy. In a blinded reader study, two radiologists rated 75\%/70\% of AI-generated reports clinically acceptable (vs.\ 95\%/100\% for originals), while our automated LLM-based judge (LLM-Eval, 61.8\%) applies a stricter standard, validating it as a conservative proxy. To our knowledge, this is the first end-to-end LI-RADS-structured reporting system for 3D liver MRI. The
code is available at \href{https://github.com/Alena-Xinran/MRI2Rep.git}{https://github.com/Alena-Xinran/MRI2Rep.git}

\end{abstract}
\section{Introduction}
\label{sec:introduction}

Radiology reports serve as the primary communication between radiologists and clinicians, translating complex imaging findings into actionable clinical insights. For 3D MRI studies, manual interpretation and report writing is time-intensive and can take radiologists 10--15 minutes per case~\cite{bhargavan2009workload}. Such per-case reporting time directly constrains clinical throughput and is associated with substantial burnout among radiologists (reported prevalence up to 49\%)~\cite{chetlen2019addressing}. While recent 3D-MRI vision--language models target representation learning, classification, or VQA~\cite{wang2026triad,su2025deciphermr,song2025livervlm,liu2025hepapathgpt}, \emph{end-to-end} generation of clinically structured, guideline-conformant reports for 3D liver MRI remains largely unaddressed.

The key challenges are twofold. First, report-level supervision is noisy: terminology varies and negation is common (e.g., ``no evidence of HCC'' implies a negative label)~\cite{irvin2019chexpert,peng2018negbio}---tolerable in large 2D datasets~\cite{johnson2019mimic} but a major overfitting risk for much smaller 3D MRI cohorts. Second, 3D MRI adds substantial computational and representational complexity: models must encode high-dimensional volumes and capture subtle cross-sequence cues while modelling long-range 3D spatial relationships under tight memory and compute budgets.

To address these challenges, we propose MRI2Rep, an autoregressive framework that directly translates 3D MRI volumes into structured radiology reports. MRI2Rep first converts noisy free-text reports into clean, closed-vocabulary supervision via a Report-to-Label Canonicalization (RLC) module (\S\ref{subsec:RLC}), then trains an encoder--decoder model (\S\ref{subsec:Autoregressive}) to predict structured diagnostic sequences from volumetric inputs, and finally renders the predicted sequences into professional reports via deterministic templates (\S\ref{subsec:rendering}). Crucially, MRI2Rep requires no lesion-level spatial annotations (bounding boxes or segmentation masks), which are expensive and rarely available in routine clinical data; lesion location is instead inferred purely from volumetric features and report-derived supervision, making the framework directly applicable to retrospective clinical cohorts.
We validate MRI2Rep on unseen test data using both automated metrics and a blinded clinical reader study, with full results reported in \S\ref{sec:Results}. Our main contributions are:
\begin{itemize}
  \item A dataset of 3{,}929 liver MRI report--image pairs, of which 3{,}830 include complete multi-phase sequences with canonicalized RLC labels.
  \item RLC, a LI-RADS-guided canonicalization module that converts free-text reports into auditable, closed-vocabulary sequences by resolving negation, uncertainty, and phrasing variation.
  \item An encoder--decoder model that autoregressively predicts structured diagnostic sequences from multi-phase 3D MRI and renders them into standardized reports.
\end{itemize}
\section{Related Work}
\label{sec:related_work}

\noindent\textbf{Medical Report Generation.}\quad
Early work focused on 2D chest X-rays using MIMIC-CXR~\cite{johnson2019mimic} and CheXpert~\cite{irvin2019chexpert} with encoder--decoder architectures~\cite{jing2018automatic,chen2020generating}. More recent 2D methods inject structured findings as an auxiliary task or report basis (ORGAN~\cite{hou2023organ}, structured multi-task learning~\cite{liang2025multitask}) and provide grounded datasets (PadChest-GR~\cite{castro2024padchestgr}), but remain 2D without LI-RADS reasoning. CT and 3D vision--language models (Merlin~\cite{blankemeier2024merlin}, CT2Rep~\cite{hamamci2024ct2rep}, M3D~\cite{bai2024m3d}, RadFM~\cite{wu2025radfm}, Triad~\cite{wang2026triad}, Decipher-MR~\cite{su2025deciphermr}) target generic reporting, representation learning, or classification; liver-MRI models address lesion classification or VQA (Liver-VLM~\cite{song2025livervlm}, HepaPathGPT~\cite{liu2025hepapathgpt}). RadGPT~\cite{bassi2025radgpt} generates reports from 3D tumour masks but needs lesion-level annotations unavailable in routine MRI. None enforce LI-RADS decision rules or produce closed-vocabulary structured output without spatial annotations.

\noindent\textbf{LI-RADS and Structured Liver Imaging.}\quad
LI-RADS~\cite{elayadi2018lirads} provides a standardized algorithm for categorizing liver observations. Prior computational work focuses on classification of pre-segmented lesions~\cite{LLD-MMRI,MedSAM2} rather than end-to-end report generation. MRI2Rep is the first to embed LI-RADS reasoning into both supervision and output vocabulary, using 9 Couinaud sectors~\cite{germann2019liver} without lesion-level annotations.

\noindent\textbf{Structured Label Extraction.}\quad
Negation, uncertainty, and phrasing variation complicate label extraction~\cite{peng2018negbio,irvin2019chexpert}. Rule-based systems address this for chest radiology; LLM-based chain-of-thought approaches~\cite{wei2022chain,kojima2022large} improve accuracy. RLC extends this to liver MRI with LI-RADS decision logic.

\section{\method}
\label{sec:Method}
\begin{figure*}[t]
    \centering
    \includegraphics[width=\linewidth]{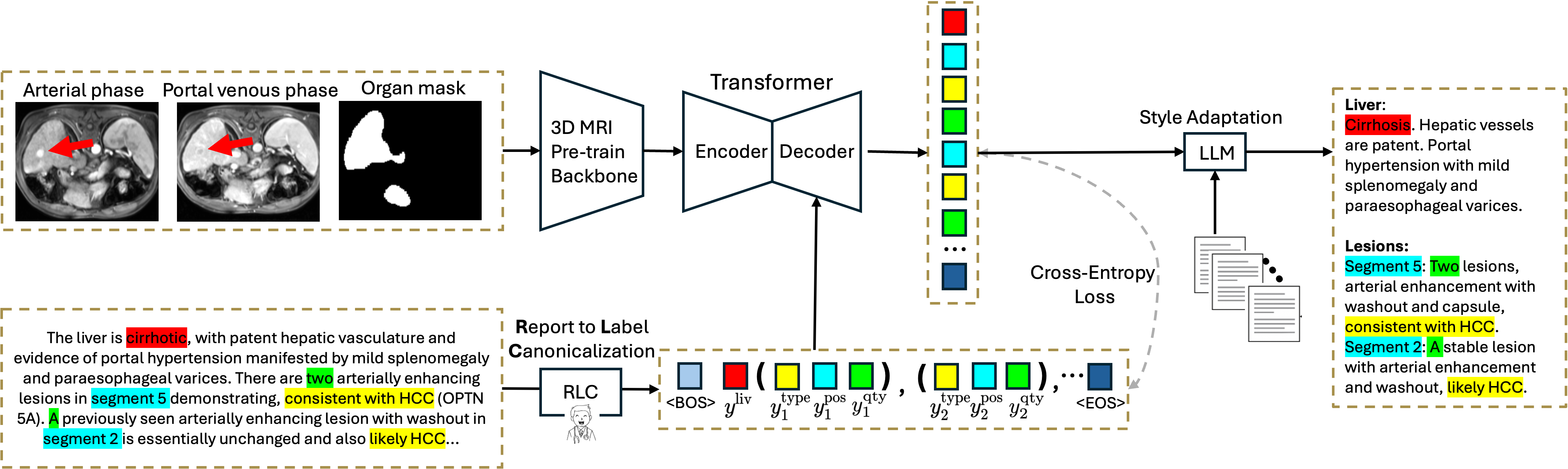}
    \caption{Overview of MRI2Rep: given a 3D liver MRI volume, the model autoregressively generates a structured radiology report by predicting a sequence of diagnostic findings directly from the image. An auxiliary classification head (detailed in \S\ref{subsec:Autoregressive}) provides an additional lesion-level supervision signal to encourage the visual encoder to retain discriminative features.}
    \label{fig:model}
\end{figure*}

MRI2Rep operates in three stages---RLC canonicalization (\S\ref{subsec:RLC}), autoregressive encoder--decoder prediction (\S\ref{subsec:Autoregressive}), and deterministic template rendering (\S\ref{subsec:rendering}); Figure~\ref{fig:model} summarizes the pipeline.

\subsection{Report-to-Label Canonicalization (RLC)} \label{subsec:RLC}

Free-text reports exhibit heterogeneous phrasing, negation, and uncertainty~\cite{peng2018negbio,irvin2019chexpert}. RLC converts each report $R$ into a structured, closed-vocabulary sequence $\mathbf{y}=(y_1,\dots,y_T)$ comprising (i) a liver-background token $y^{\mathrm{liv}} \in \mathcal{Y}_{\mathrm{liv}}$ and (ii) up to $K=8$ lesion triplets $\{(y^{\mathrm{type}}_k, y^{\mathrm{pos}}_k, y^{\mathrm{qty}}_k)\}$, serialized as \texttt{[BOS, liver, type$_1$, pos$_1$, qty$_1$, \dots, EOS]}; each triplet is paired with an evidence sentence $e_k \subset R$ for auditability~\cite{jain2021radgraph}.

\begin{figure*}[t]
    \centering
\includegraphics[width=\linewidth]{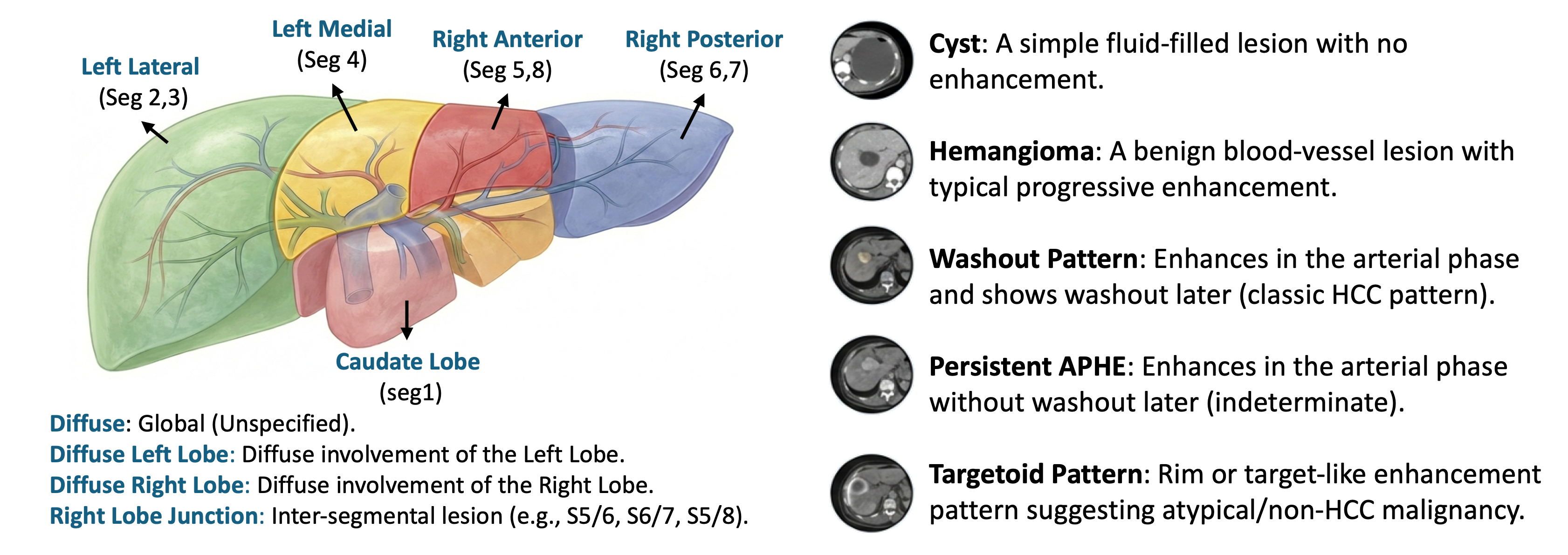}
    \caption{Output vocabulary: location sectors $\mathcal{Y}_{\mathrm{pos}}$ (left, Couinaud-derived) and lesion types $\mathcal{Y}_{\mathrm{type}}$ with representative MRI examples (right).}
    \label{fig:label}
\end{figure*}

Concretely, RLC is an LI-RADS-guided LLM prompting protocol, not a black-box call: an LLM (Claude-3.5; GPT-5 and DeepSeek-R1 also evaluated, Table~\ref{tab:retention_llm_t1t2}) receives the LI-RADS major-feature decision logic with chain-of-thought instructions~\cite{wei2022chain,kojima2022large} and, per candidate observation, emits a $(y^{\mathrm{type}}_k, y^{\mathrm{pos}}_k, y^{\mathrm{qty}}_k)$ triplet \emph{with} the verbatim evidence sentence $e_k\subset R$ licensing it; triplets are assembled deterministically into $\mathbf{y}$ and abstained when evidence is absent, making every label reproducible and auditable. Following the LI-RADS algorithm~\cite{elayadi2018lirads}, the prompt queries sequentially for definitive benign features, targetoid features, and washout dynamics to assign lesion categories. The four token subsets are: $\mathcal{Y}_{\mathrm{liv}}$ (2 tokens: normal, fibrotic/cirrhotic); $\mathcal{Y}_{\mathrm{type}}$ (5 LI-RADS categories), $\mathcal{Y}_{\mathrm{pos}}$ (9 Couinaud-derived surgical sectors~\cite{germann2019liver}), and $\mathcal{Y}_{\mathrm{qty}}$ (4 quantity buckets: single, 2, $\geq$3, diffuse); representative examples of $\mathcal{Y}_{\mathrm{type}}$ and $\mathcal{Y}_{\mathrm{pos}}$ are shown in Fig.~\ref{fig:label}. The shared token dictionary $\mathcal{V}$ contains special tokens, liver-background tokens, lesion-type tokens, location tokens, quantity-bucket tokens, and \texttt{NO\_LESION}, giving $|\mathcal{V}|=24$. Although the LI-RADS categories (LR-1--5) formally apply only to patients at risk for HCC, our vocabulary additionally includes non-HCC tokens (cyst, hemangioma, targetoid) and normal/cirrhotic liver-background tokens, extending coverage beyond the at-risk population.

\subsection{Encoder–Decoder Architecture} \label{subsec:Autoregressive}
Following 3D radiology-captioning practice~\cite{hamamci2024ct2rep,blankemeier2024merlin}, we model report generation as an autoregressive conditional distribution $p(\mathbf{y}\mid \mathbf{V})=\prod_{t=1}^{T} p(y_t \mid y_{<t}, \mathbf{V})$, where $\mathbf{V} \in \mathbb{R}^{3 \times H \times W \times D}$ stacks the arterial-phase (ART), portal-venous-phase (PV), and a binary liver segmentation mask (each independently intensity-normalised).

The visual encoder $f_{\theta}$ is a 3D CNN with four Conv3d-IN-GELU blocks and stride-2 downsampling, producing visual memory tokens $\mathbf{E}^{\mathrm{vis}} = f_{\theta}(\mathbf{V}) \in \mathbb{R}^{S_{\mathrm{vis}} \times d}$. A visual Transformer applies self-attention for global 3D context; the autoregressive decoder $g_{\phi}$~\cite{vaswani2017attention} then cross-attends to $\mathbf{E}^{\mathrm{vis}}$ and projects to logits over $\mathcal{V}$.

The backbone is pre-trained to reconstruct the PV phase from the ART input (MSE and SSIM loss within the liver mask) before joint fine-tuning with the Transformer. To make the decoder attend to visual features rather than text co-occurrence, we add two mechanisms: (i) an auxiliary binary head---global average pooling over $\mathbf{E}^{\mathrm{vis}}$ feeds a linear classifier predicting lesion presence $\hat{z}\in\{0,1\}$, encouraging the backbone to retain lesion-discriminative features; and (ii) word dropout~\cite{iyyer2015deep}, randomly replacing a fraction of decoder input tokens with \texttt{PAD} to reduce reliance on the text prefix.
The training objective is:
\begin{equation}
  \mathcal{L} = \mathcal{L}_{\mathrm{CE}}(\mathbf{y}, \hat{\mathbf{y}}) + \lambda\,\mathcal{L}_{\mathrm{BCE}}(z, \hat{z}),
\end{equation}
where $\mathcal{L}_{\mathrm{CE}}$ uses per-class weights~\cite{cui2019classbalanced} and label smoothing.

\noindent\textbf{Implementation.} Volumes are resized to $192\!\times\!192\!\times\!96$; ART/PV normalised independently. CNN: four Conv3d-IN-GELU blocks (channels 32/64/128/256; strides 2/2/2/1), yielding 256-d visual tokens. Decoder: 6 layers, 8 heads, FFN 512, dropout 0.3. AdamW (lr $5\!\times\!10^{-5}$), cosine schedule, batch 12, early stopping on val Lesion-F1. NVIDIA A6000/RTX 4090.

\subsection{Sequence-to-Report Rendering} \label{subsec:rendering}
The predicted token sequence is deterministically mapped to a natural-language report via hand-crafted sentence templates: the liver-background token selects a parenchymal description and each lesion triplet yields a finding sentence (e.g., ``A single observation consistent with definite HCC is identified in the right posterior sector''). Stylistic variation is introduced by sampling from a small pool of paraphrase alternatives per template, produced by prompting Claude-3.5 with 2,000 randomly sampled training-set reports to capture institutional writing conventions; this prevents fixed lexical artefacts from unblinding readers in the reader study (\S\ref{subsec:reader_study}).

\section{Results}
\label{sec:Results}

\noindent\textbf{Dataset \& evaluation setup.}\quad
The cohort comprises 3{,}929 multi-phase liver-MRI report--image pairs from a 10-year retrospective collection at a single tertiary academic hospital, acquired on Siemens scanners (predominantly 3T) with extracellular gadolinium-based contrast; reports were written by board-certified abdominal radiologists. After RLC canonicalization (\S\ref{subsec:RLC}), the 3{,}830 studies with complete arterial and portal-venous phases are split 80/10/10\% (train/val/test) by a stratified set-cover on (type,\,pos,\,qty), yielding 383 held-out test cases that cover rare LI-RADS categories. $k$-fold cross-validation is impractical for this 3D autoregressive pipeline; the set-cover split mitigates split variance, and the $\geq$21-point Lesion-F1 margin over the best baseline ($n{=}383$) lies well outside any plausible bootstrap interval.

\subsection{RLC Validation}
\begin{table*}[t]
\centering
\caption{Radiologist retention rate (\% Retained = Rad/AI) across LLM backends for labels extracted from liver MRI radiology reports.
Gray parentheses report the underlying counts (Retained\,/\,Produced).}
\label{tab:retention_llm_t1t2}

\setlength{\tabcolsep}{5pt}
\renewcommand{\arraystretch}{1.2}

\footnotesize
\begin{tabular*}{\textwidth}{@{\extracolsep{\fill}}lcccc}
\toprule
\multirow{2}{*}{} & \multirow{2}{*}{RLC}
& \multicolumn{3}{c}{LLM backend} \\
\cmidrule(lr){3-5}
& & Claude-3.5~\cite{anthropic_claude35sonnet_announcement_2024} & GPT-5~\cite{singh2025openai} & DeepSeek-R1~\cite{guo2025deepseekr1_nature} \\
\midrule

\multirow{2}{*}{Organ (Liver)}
& $\times$   
& 76.1 {\textcolor{gray}{\scriptsize(1619/2128)}} 
& 84.6 {\textcolor{gray}{\scriptsize(1801/2128)}} 
& 80.1 {\textcolor{gray}{\scriptsize(1705/2128)}} \\
& $\checkmark$
& 97.1 {\textcolor{gray}{\scriptsize(1652/1701)}} 
& \textbf{97.7} {\textcolor{gray}{\scriptsize(1662/1701)}} 
& 95.2 {\textcolor{gray}{\scriptsize(1620/1701)}} \\
\midrule

\multirow{2}{*}{Lesion (Non-HCC)}
& $\times$   
& 82.3 {\textcolor{gray}{\scriptsize(1455/1769)}} 
& 84.2 {\textcolor{gray}{\scriptsize(1490/1769)}} 
& 80.7 {\textcolor{gray}{\scriptsize(1428/1769)}} \\
& $\checkmark$
& \textbf{98.0} {\textcolor{gray}{\scriptsize(1498/1528)}} 
& 96.3 {\textcolor{gray}{\scriptsize(1472/1528)}} 
& 92.3 {\textcolor{gray}{\scriptsize(1411/1528)}} \\
\midrule

\multirow{2}{*}{Lesion (HCC)}
& $\times$   
& 77.9 {\textcolor{gray}{\scriptsize(392/503)}} 
& 82.7 {\textcolor{gray}{\scriptsize(416/503)}} 
& 76.7 {\textcolor{gray}{\scriptsize(386/503)}} \\
& $\checkmark$
& \textbf{98.1} {\textcolor{gray}{\scriptsize(418/426)}} 
& 97.0 {\textcolor{gray}{\scriptsize(413/426)}} 
& 91.8 {\textcolor{gray}{\scriptsize(391/426)}} \\
\bottomrule
\end{tabular*}
\normalsize
\end{table*}
We assess RLC reliability via a radiologist retention test on each category's reports: two radiologists (junior, senior) independently mark each AI-produced label as retained if supported by the original report (and evidence sentence when available), averaged across annotators. Table~\ref{tab:retention_llm_t1t2} compares direct taxonomy mapping (RLC disabled) against evidence-backed canonicalization (RLC enabled) across three LLM backends: RLC consistently raises retention from $\sim$77--84\% to $\geq$92\% across all categories and backends, with produced labels slightly reduced as RLC abstains when evidence is insufficient.

\subsection{Ablation Study}

Predicted lesions are scored on the held-out test set by matching \emph{type} and \emph{location} (and optionally \emph{quantity}) to the ground truth. We report: case-level \textbf{Sen/Spe} (whether a case contains any lesion); \textbf{Lesion-F1}, set-F1 over matched (type,\,pos) pairs on lesion-positive cases; \textbf{Triplet-F1}, strict set-F1 over (type, pos, qty); \textbf{Liver-Acc} for the liver-background token; and \textbf{LLM-Eval}, the fraction of rendered reports rated acceptable (score 1--2) by Claude-3.5~\cite{anthropic_claude35sonnet_announcement_2024} on a 3-point scale, given rendered and reference reports in randomised order.

\begin{table}[t]
\centering
\caption{\textbf{Ablation and comparison study.}
We report lesion \emph{detection} (Sen/Spe), lesion \emph{characterisation} as set-F1 over (type,\,pos) pairs (Lesion-F1) and strict (type,\,pos,\,qty) triplets (Triplet-F1), organ-context classification (Liver-Acc), and Claude-3.5~\cite{anthropic_claude35sonnet_announcement_2024} judged clinical consistency (LLM-Eval; see \S\ref{sec:Results} for details).
All values are in \%; \textbf{bold} indicates the best result.}
\label{tab:lesion_liver_metrics}

\setlength{\tabcolsep}{3pt}
\renewcommand{\arraystretch}{1.15}
\footnotesize
\begin{tabular*}{\textwidth}{@{\extracolsep{\fill}}lcccccc}
\toprule
Method & Sen$\uparrow$ & Spe$\uparrow$ & Lesion-F1$\uparrow$ & Triplet-F1$\uparrow$ & Liver-Acc$\uparrow$ & LLM-Eval$\uparrow$ \\
\midrule
\multicolumn{7}{l}{\textit{Comparison with prior work}} \\[2pt]
M3D~\cite{bai2024m3d}               & 10.5 & 86.3          & 6.4 & 4.1 & 58.2 & 17.8 \\
CT2Rep~\cite{hamamci2024ct2rep}     & 5.2  & 91.8          & 7.2 & 4.8 & 61.5 & 14.6 \\
Merlin~\cite{blankemeier2024merlin} & 25.1 & 84.7          & 5.8 & 3.9 & 55.6 & 21.3 \\
RadFM~\cite{wu2025radfm}            & 21.6 & \textbf{92.4} & 8.3 & 5.6 & 66.8 & 25.2 \\
\midrule
\multicolumn{7}{l}{\textit{Architecture baseline}} \\[2pt]
Multi-label cls.  & 43.2 & 74.1 & 14.8 & 9.3 & 79.6 & 33.4 \\
\midrule
\multicolumn{7}{l}{\textit{Ablation}} \\[2pt]
w/o Vis. Enc                        & 42.7 & 60.9 &10.4 & 5.9 & 76.8 & 29.7 \\
w/o Pretrain                       & 57.3 & 50.6 & 22.1  & 21.3 & 79.1 & 47.9 \\
w/o PV                              & 46.4 & 77.8 & 13.5  & 9.9  & 77.8 & 37.1 \\
\textbf{MRI2Rep (Ours)}             & \textbf{76.0} & 68.4 & \textbf{29.4} & \textbf{28.1} & \textbf{82.4} & \textbf{61.8} \\
\bottomrule
\end{tabular*}
\normalsize
\end{table}

Table~\ref{tab:lesion_liver_metrics} compares MRI2Rep with 3D vision--language baselines and ablates key choices. For fair comparison, baseline free-text outputs are mapped to the same closed (type,\,pos,\,qty) taxonomy and liver-background tokens via Claude-3.5~\cite{anthropic_claude35sonnet_announcement_2024}.

MRI2Rep achieves the highest sensitivity (76.0\%) and Lesion-F1 (29.4\%); label quality is independently validated by the 98.1\% retention rate (Table~\ref{tab:retention_llm_t1t2}). Specificity (68.4\%) reflects a deliberate sensitivity--specificity trade-off (oversampling, class-reweighted loss) tunable at inference; prioritising sensitivity aligns with guidelines favouring minimal missed malignancies. The \emph{Multi-label cls.}\ baseline's lower Lesion-F1 (14.8\%) confirms autoregressive decoding better captures inter-lesion dependencies, and the narrow Lesion-F1/Triplet-F1 gap (29.4 vs.\ 28.1) indicates reliable quantity prediction once a lesion is localised.

Removing ART-to-PV pre-training (\emph{w/o Pretrain}) drops Lesion-F1 to 22.1 (Sen 57.3), confirming pre-training as essential initialisation. Dropping the portal-venous phase (\emph{w/o PV}) causes the largest fall (Lesion-F1 13.5; Sen 46.4): PV washout dynamics are essential for HCC and unrecoverable from ART alone. Removing visual self-attention (\emph{w/o Vis.\ Enc.}) lowers Lesion-F1 to 10.4 and Spe to 60.9\%, as the decoder over-generates lesions without global 3D context.

\subsection{Clinical Reader Study}
\label{subsec:reader_study}

To validate LLM-Eval, two radiologists (junior/senior) independently scored 100 randomly sampled, randomly ordered test cases (40 AI-generated, 60 original controls) given only the MRI images, blinded to report origin and to the AI:control ratio, on a 3-point scale (acceptance = score 1--2). Radiologists reported no systematic strategy for identifying AI reports, supporting the integrity of the blind.

Radiologists accepted 75\%/70\% of AI reports vs.\ 95\%/100\% for originals, with almost-perfect agreement (Cohen's $\kappa=0.97$, weighted $0.96$). LLM-Eval's lower rate (61.8\%) shows Claude-3.5 applies a \emph{stricter} standard than human readers---a desirable conservative bias---and, together with its $\geq$98\% label-extraction accuracy (Table~\ref{tab:retention_llm_t1t2}), validates it as a reliable, high-throughput automated judge.

\subsection{Case Study}

\begin{figure*}[t]
    \centering
    \includegraphics[width=\linewidth]{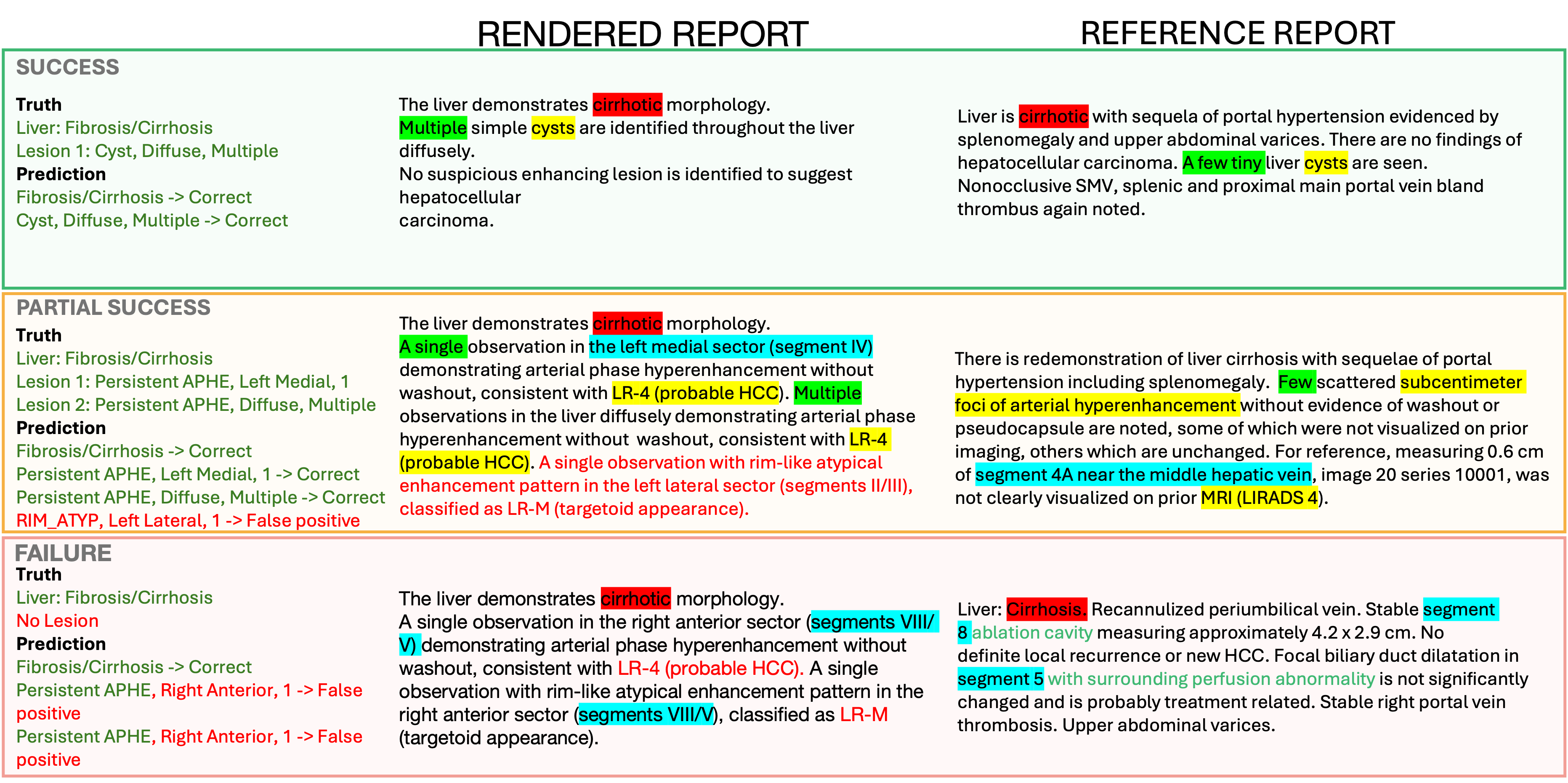}
    \caption{\textbf{Qualitative examples} from the held-out test set. Each row shows the ground-truth and predicted structured labels (left), the rendered report (centre), and the reference report (right), with key clinical terms highlighted. 
    (a)~Full success: all predicted labels match the ground truth; the rendered report captures the essential diagnostic content.
    (b)~Partial success: liver background and two lesions are correctly predicted, but one false-positive LR-M is hallucinated.
    (c)~Failure: the model predicts two active lesions where the reference describes only post-ablation changes, with no evidence of HCC recurrence.}
    \label{fig:qualitative}
\end{figure*}

Three representative cases (Fig.~\ref{fig:qualitative}) span a full success (a), a partial success with one hallucinated LR-M (b), and a failure where post-ablation changes are misread as two active lesions (c), illustrating the difficulty of complex multi-finding sequences.

Systematic error analysis reveals three patterns. Per-class Lesion-F1 ranges from 51.7\% (cysts) to 11.3\% (targetoid, subtle rim enhancement), with washout-pattern lesions (LR-4/5) at 24.6\%, reflecting hard cross-phase reasoning. Lesion-F1 drops from 38.4\% (single-lesion) to 21.7\% (multi-lesion, $\geq$2), as longer sequences accumulate inter-triplet errors. Of all false positives, 34.1\% occur in prior-ablation cases where residual enhancement mimics active lesions, motivating treatment history as a future input.

\section{Conclusion}

MRI2Rep is an autoregressive framework for structured liver MRI report generation: its LI-RADS-guided RLC module yields clean, closed-vocabulary supervision without lesion-level annotations, enabling an encoder--decoder model that outperforms medical vision--language baselines and reaches 75\%/70\% blinded clinical acceptability. Key limitations: report-only supervision without spatial annotations limits lesion-level F1 and multi-lesion/post-treatment cases (a dominant false-positive source, \S\ref{sec:Results}); the closed RLC vocabulary bounds expressible semantics; and single-centre data leaves cross-centre generalisation open. Future work includes treatment-history integration, lesion-level spatial supervision, and external multi-site validation.

\begin{credits}
\subsubsection{\ackname} This work was supported by the National Institutes of Health (NIH) under Award Number R01EB037669.

\subsubsection{\discintname}
The authors declare no competing interests.
\end{credits}

\bibliographystyle{splncs04}
\bibliography{ref}

@article{bhargavan2009workload,
  title={Workload of radiologists in United States in 2006--2007 and trends since 1991--1992},
  author={Bhargavan, Mythreyi and Kaye, Alan H and Forman, Howard P and Sunshine, Jonathan H},
  journal={Radiology},
  volume={252},
  number={2},
  pages={458--467},
  year={2009},
  publisher={Radiological Society of North America}
}

@article{chetlen2019addressing,
  title={Addressing burnout in radiologists},
  author={Chetlen, Alicia L and Chan, Tracy L and Ballard, David H and Frigini, Lisa A and Hildebrand, Adam and Kim, Soyoung and others},
  journal={Academic Radiology},
  volume={26},
  number={4},
  pages={526--533},
  year={2019},
  publisher={Elsevier}
}

@article{johnson2019mimic,
  title={MIMIC-CXR, a de-identified publicly available database of chest radiographs with free-text reports},
  author={Johnson, Alistair EW and Pollard, Tom J and Berkowitz, Seth J and Greenbaum, Nathaniel R and Lungren, Matthew P and Deng, Chih-ying and Mark, Roger G and Horng, Steven},
  journal={Scientific data},
  volume={6},
  number={1},
  pages={317},
  year={2019},
  publisher={Nature Publishing Group UK London}
}

@inproceedings{irvin2019chexpert,
  title={Chexpert: A large chest radiograph dataset with uncertainty labels and expert comparison},
  author={Irvin, Jeremy and Rajpurkar, Pranav and Ko, Michael and Yu, Yifan and Ciurea-Ilcus, Silviana and Chute, Chris and Marklund, Henrik and Haghgoo, Behzad and Ball, Robyn and Shpanskaya, Katie and others},
  booktitle={Proceedings of the AAAI conference on artificial intelligence},
  volume={33},
  pages={590--597},
  year={2019}
}

@article{peng2018negbio,
  title={NegBio: a high-performance tool for negation and uncertainty detection in radiology reports},
  author={Peng, Yifan and Wang, Xiaosong and Lu, Le and Bagheri, Mohammadhadi and Summers, Ronald and Lu, Zhiyong},
  journal={AMIA Summits on Translational Science Proceedings},
  volume={2018},
  pages={188},
  year={2018}
}

@article{wei2022chain,
  title={Chain-of-thought prompting elicits reasoning in large language models},
  author={Wei, Jason and Wang, Xuezhi and Schuurmans, Dale and Bosma, Maarten and Xia, Fei and Chi, Ed and Le, Quoc V and Zhou, Denny and others},
  journal={Advances in neural information processing systems},
  volume={35},
  pages={24824--24837},
  year={2022}
}

@inproceedings{vaswani2017attention,
  title={Attention Is All You Need},
  author={Vaswani, Ashish and Shazeer, Noam and Parmar, Niki and Uszkoreit, Jakob and Jones, Llion and Gomez, Aidan N and Kaiser, {\L}ukasz and Polosukhin, Illia},
  booktitle={Advances in Neural Information Processing Systems (NeurIPS)},
  volume={30},
  year={2017}
}

@inproceedings{cui2019classbalanced,
  title={Class-Balanced Loss Based on Effective Number of Samples},
  author={Cui, Yin and Jia, Menglin and Lin, Tsung-Yi and Song, Yang and Belongie, Serge},
  booktitle={Proceedings of the IEEE/CVF Conference on Computer Vision and Pattern Recognition (CVPR)},
  year={2019}
}

@inproceedings{iyyer2015deep,
  title={Deep Unordered Composition Rivals Syntactic Methods for Text Classification},
  author={Iyyer, Mohit and Manjunatha, Varun and Boyd-Graber, Jordan and Daum{\'e} III, Hal},
  booktitle={Proceedings of the 53rd Annual Meeting of the Association for Computational Linguistics (ACL)},
  year={2015}
}

@article{bai2024m3d,
  title={M3d: Advancing 3d medical image analysis with multi-modal large language models},
  author={Bai, Fan and Du, Yuxin and Huang, Tiejun and Meng, Max Q-H and Zhao, Bo},
  journal={arXiv preprint arXiv:2404.00578},
  year={2024}
}

@inproceedings{hamamci2024ct2rep,
  title={CT2Rep: Automated Radiology Report Generation for 3D Medical Imaging},
  author={Hamamci, Ibrahim Ethem and others},
  booktitle={International Conference on Medical Image Computing and Computer-Assisted Intervention (MICCAI)},
  year={2024},
  organization={Springer}
}

@inproceedings{bassi2025radgpt,
  title={Radgpt: Constructing 3d image-text tumor datasets},
  author={Bassi, Pedro RAS and Yavuz, Mehmet Can and Hamamci, Ibrahim Ethem and Er, Sezgin and Chen, Xiaoxi and Li, Wenxuan and Menze, Bjoern and Decherchi, Sergio and Cavalli, Andrea and Wang, Kang and others},
  booktitle={Proceedings of the IEEE/CVF International Conference on Computer Vision},
  pages={23720--23730},
  year={2025}
}

@misc{anthropic_claude35sonnet_announcement_2024,
  title        = {Claude 3.5 Sonnet},
  author       = {{Anthropic}},
  year         = {2024},
  note         = {Announcements, Jun 21, 2024},
  url          = {https://www.anthropic.com/news/claude-3-5-sonnet}
}

@misc{singh2025openai,
  title  = {{GPT-5} System Card},
  author = {{OpenAI}},
  year   = {2025},
  howpublished = {OpenAI technical report},
  url    = {https://openai.com/index/gpt-5-system-card/}
}

@article{guo2025deepseekr1_nature,
  title   = {DeepSeek-R1 incentivizes reasoning in LLMs through reinforcement learning},
  author  = {Guo, Daya and Yang, Dejian and Zhang, Haowei and others},
  journal = {Nature},
  volume  = {645},
  pages   = {633--638},
  year    = {2025},
  doi     = {10.1038/s41586-025-09422-z}
}

@inproceedings{jain2021radgraph,
  title     = {{RadGraph}: Extracting Clinical Entities and Relations from Radiology Reports},
  author    = {Jain, Saahil and Agrawal, Ashwin and Saporta, Adriel and Truong, Steven and Duong, Du Nguyen and Bui, Tan and Chambon, Pierre and Zhang, Yuhao and Lungren, Matthew P. and Ng, Andrew Y. and Langlotz, Curtis P. and Rajpurkar, Pranav},
  booktitle = {Proceedings of the Neural Information Processing Systems Track on Datasets and Benchmarks (NeurIPS Datasets and Benchmarks)},
  volume    = {1},
  year      = {2021}
}

@misc{blankemeier2024merlin,
  title         = {Merlin: A Vision Language Foundation Model for 3D Computed Tomography},
  author        = {Blankemeier, Louis and Cohen, Joseph Paul and Kumar, Ashwin and Van Veen, Dave and Gardezi, Syed Jamal Safdar and Paschali, Magdalini and Chen, Zhihong and Delbrouck, Jean-Benoit and Reis, Eduardo and Truyts, Cesar and Bluethgen, Christian and Jensen, Malte Engmann Kjeldskov and Ostmeier, Sophie and Varma, Maya and Valanarasu, Jeya Maria Jose and Fang, Zhongnan and Huo, Zepeng and Nabulsi, Zaid and Ardila, Diego and Weng, Wei-Hung and Amaro Junior, Edson and Ahuja, Neera and Fries, Jason and Shah, Nigam H. and Johnston, Andrew and Boutin, Robert D. and Wentland, Andrew and Langlotz, Curtis P. and Hom, Jason and Gatidis, Sergios and Chaudhari, Akshay S.},
  year          = {2024},
  eprint        = {2406.06512},
  archivePrefix = {arXiv},
  primaryClass  = {cs.CV},
  doi           = {10.48550/arXiv.2406.06512}
}

@article{wu2025radfm,
  title   = {Towards generalist foundation model for radiology by leveraging web-scale 2D\&3D medical data},
  author  = {Wu, Chaoyi and Zhang, Xiaoman and Zhang, Ya and Hui, Hui and Wang, Yanfeng and Xie, Weidi},
  journal = {Nature Communications},
  year    = {2025},
  volume  = {16},
  number  = {1},
  pages   = {7866},
  doi     = {10.1038/s41467-025-62385-7}
}

@article{elayadi2018lirads,
  title={LI-RADS major features: CT, MRI with extracellular agents, and MRI with hepatobiliary agents},
  author={Santillan, Cynthia and Fowler, Kathryn and Kono, Yuko and Chernyak, Victoria},
  journal={Abdominal Radiology},
  volume={43},
  number={1},
  pages={75--81},
  year={2018},
  publisher={Springer},
  doi={10.1007/s00261-017-1291-4}
}

@article{germann2019liver,
  title={Liver segmentation: practical tips},
  author={Germain, Thibaut and Favelier, Sylvain and Cercueil, Jean-Pierre and Denys, Alban and Krause, Denis and Guiu, Boris},
  journal={Diagnostic and Interventional Imaging},
  volume={95},
  number={11},
  pages={1003--1016},
  year={2014},
  publisher={Elsevier},
  doi={10.1016/j.diii.2013.12.005}
}

@inproceedings{kojima2022large,
  title={Large Language Models are Zero-Shot Reasoners},
  author={Kojima, Takeshi and Gu, Shixiang Shane and Reid, Machel and Matsuo, Yutaka and Iwasawa, Yusuke},
  booktitle={Advances in Neural Information Processing Systems},
  volume={35},
  pages={22199--22213},
  year={2022}
}

@inproceedings{jing2018automatic,
  title     = {On the Automatic Generation of Medical Imaging Reports},
  author    = {Jing, Baoyu and Xie, Pengtao and Xing, Eric},
  booktitle = {Proceedings of the 56th Annual Meeting of the Association
               for Computational Linguistics (Volume 1: Long Papers)},
  pages     = {2577--2586},
  year      = {2018},
  address   = {Melbourne, Australia},
  publisher = {Association for Computational Linguistics},
  doi       = {10.18653/v1/P18-1240}
}

@inproceedings{chen2020generating,
  title={Generating radiology reports via memory-driven transformer},
  author={Chen, Zhihong and Song, Yan and Chang, Tsung-Hui and Wan, Xiang},
  booktitle={Proceedings of the 2020 conference on empirical methods in natural language processing (EMNLP)},
  pages={1439--1449},
  year={2020}
}

@article{LLD-MMRI,
  title={Sdr-former: A siamese dual-resolution transformer for liver lesion classification using 3d multi-phase imaging},
  author={Lou, Meng and Ying, Hanning and Liu, Xiaoqing and Zhou, Hong-Yu and Zhang, Yuqin and Yu, Yizhou},
  journal={Neural Networks},
  pages={107228},
  year={2025}
}

@article{MedSAM2,
    title={MedSAM2: Segment Anything in 3D Medical Images and Videos},
    author={Ma, Jun and Yang, Zongxin and Kim, Sumin and Chen, Bihui and Baharoon, Mohammed and Fallahpour, Adibvafa and Asakereh, Reza and Lyu, Hongwei and Wang, Bo},
    journal={arXiv preprint arXiv:2504.03600},
    year={2025}
}

@article{liu2025hepapathgpt,
  title   = {A generative vision-language model for holistic pathological assessment using preoperative imaging in hepatocellular carcinoma},
  author  = {Wang, Liyang and others},
  journal = {eBioMedicine},
  volume  = {122},
  pages   = {106060},
  year    = {2025},
  publisher = {Elsevier},
  doi     = {10.1016/j.ebiom.2025.106060}
}

@article{song2025livervlm,
  title   = {Liver-VLM: Enhancing Focal Liver Lesion Classification with Self-Supervised Vision-Language Pretraining},
  author  = {Song, Jian and Hu, Yi and Wang, Hao and Chen, Yen-Wei},
  journal = {Applied Sciences},
  volume  = {15},
  number  = {23},
  pages   = {12578},
  year    = {2025},
  publisher = {MDPI},
  doi     = {10.3390/app152312578}
}

@article{wang2026triad,
  title   = {Vision foundation model for 3D magnetic resonance imaging segmentation, classification, and registration},
  author  = {Wang, Shansong and Safari, Mojtaba and Li, Qiang and Chang, Chih-Wei and Qiu, Richard L. J. and Roper, Justin and Yu, David S. and Yang, Xiaofeng},
  journal = {Medical Image Analysis},
  volume  = {110},
  pages   = {103992},
  year    = {2026},
  publisher = {Elsevier},
  doi     = {10.1016/j.media.2026.103992}
}

@article{su2025deciphermr,
  title         = {Decipher-MR: A Vision-Language Foundation Model for 3D MRI Representations},
  author        = {Yang, Zhijian and DSouza, Noel and Megyeri, Istvan and others},
  journal       = {arXiv preprint arXiv:2509.21249},
  year          = {2025}
}

@article{liang2025multitask,
  title   = {Multi-Task Learning for Radiology Report Generation with Structured Findings Consistency},
  author  = {Liang, Jianguo},
  journal = {Computer Science Bulletin},
  volume  = {8},
  number  = {1},
  pages   = {477--489},
  year    = {2025},
  doi     = {10.71465/csb178}
}

@article{hou2023organ,
  title   = {ORGAN: Observation-Guided Radiology Report Generation via Tree Reasoning},
  author  = {Hou, Wenjun and Xu, Kaishuai and Cheng, Yi and Li, Wenjie and Liu, Jiang},
  journal = {arXiv preprint arXiv:2306.06466},
  year    = {2023}
}

@article{castro2024padchestgr,
  title   = {PadChest-GR: A Bilingual Chest X-Ray Dataset for Grounded Radiology Report Generation},
  author  = {Castro, Daniel C. and Bustos, Aurelia and Bannur, Shruthi and others},
  journal = {arXiv preprint arXiv:2411.05085},
  year    = {2024}
}
\clearpage

\end{document}